\documentclass[sigconf]{acmart}
\AtBeginDocument{%
  }
\usepackage{algorithm}
\usepackage{algpseudocode}
\usepackage{subcaption}

\setcopyright{acmlicensed}
\copyrightyear{2018}
\acmYear{2018}
\acmDOI{XXXXXXX.XXXXXXX}
\acmConference[Conference acronym 'XX]{Make sure to enter the correct
  conference title from your rights confirmation email}{June 03--05,
  2018}{Woodstock, NY}
\acmISBN{978-1-4503-XXXX-X/2018/06}

\begin{document}

\title{Targeted Label-Flipping and Oversampling Attacks on Federated Conditional GANs}

\author{Panav Shah}
\email{panav.shah@iitb.ac.in}
\affiliation{%
  \institution{Indian Institute of Technology, Bombay}
  \city{Mumbai}
  \country{India}
}

\author{Avishek Ghosh}
\email{avishek\_ghosh@iitb.ac.in}
\affiliation{%
  \institution{Indian Institute of Technology, Bombay}
  \city{Mumbai}
  \country{India}
}

\renewcommand{\shortauthors}{Shah et. al.}

\begin{abstract}
In a federated learning setup for GANs, several adversarial attacks are possible. One such attack is label flipping, in which malicious clients deliberately alter label information during local training in order to manipulate the global generator. The objective of this attack is to skew the learned generation distribution so that samples conditioned on a target label are instead mapped to a source class. In this work, we investigate the effectiveness of label-flipping attacks in federated GANs through both theoretical analysis and empirical evaluation. We further consider an oversampling-based variant, in which malicious clients upweight poisoned samples during local training to amplify their influence on the aggregated global model. We quantify the resulting distributional shift by computing the Kullback–Leibler divergence between the clean and poisoned class-conditional distributions, and show — both analytically and on FEMNIST, MNIST, and CIFAR-10 — that the semantic damage of the attack grows linearly in the effective poisoning strength while deviation from the true target distribution grows only quadratically, making the attack effective yet difficult to detect from label-agnostic metrics.
\end{abstract}






\maketitle

\section{Introduction}
Generative Adversarial Networks (GANs)~\cite{goodfellow2014generativeadversarialnetworks} learn data distributions through an adversarial game between a generator and a discriminator, while conditional GANs (cGANs)~\cite{mirza2014conditionalgenerativeadversarialnets} extend this framework by incorporating label information to enable class-conditional generation. Owing to their expressive power, cGANs are widely used in applications such as image synthesis and data augmentation.

Federated Learning (FL)~\cite{Li_2023} enables multiple clients to collaboratively train a shared model without exchanging raw data, typically via Federated Averaging (FedAvg)~\cite{mcmahan2023communicationefficientlearningdeepnetworks}. While FL improves data privacy, it also introduces new attack surfaces, as the server generally assumes client updates to be benign. In adversarial settings~\cite{goodfellow2014explaining}, malicious clients can manipulate their local training to corrupt the global model. Prior work~\cite{10024252} has studied poisoning attacks in FL, including noise injection, label flipping, oversampling, and backdoor attacks, but their impact on federated generative models~\cite{rasouli2020fedganfederatedgenerativeadversarial,Wu_2022}---particularly conditional GANs---remains relatively underexplored.

In this work, we study targeted label flipping and oversampling-based label flipping attacks in federated conditional GANs. We theoretically characterize how these attacks bias the learned class-conditional distributions under FedAvg and empirically validate our analysis on the FEMNIST~\cite{caldas2019leafbenchmarkfederatedsettings}, MNIST~\cite{lecun1998mnist} and CIFAR-10~\cite{Krizhevsky09learningmultiple} datasets. Our results show that even simple label manipulation can significantly distort conditional generation, while oversampling amplifies the attack without a proportional increase in detectability.

\paragraph{Contributions.} Our main contributions are as follows. (i) We formulate two targeted label-manipulation attacks --- \emph{simple label flipping} and \emph{oversampling-based label flipping} --- in the federated cGAN setting, with algorithmic descriptions of the adversarial client update. (ii) We derive a mixture-model characterization of the poisoned class-conditional distribution, showing the source--target KL divergence decreases \emph{linearly} in the effective poisoning strength $\beta$, while deviation from the true target distribution grows only \emph{quadratically} in $\beta$ --- an asymmetry indicating the attack is effective yet hard to detect from the target distribution alone. (iii) We empirically validate these predictions on FEMNIST, MNIST and CIFAR-10 under a realistic non-IID federated setup, finding close agreement between theory and experiment for both attack variants.
\section{Related Work}
\label{sec:related}

\paragraph{Federated learning and poisoning attacks.}
Federated Averaging (FedAvg)~\cite{mcmahan2023communicationefficientlearningdeepnetworks} is the canonical aggregation protocol in federated learning, and a substantial body of work has examined its robustness to adversarial participants~\cite{Li_2023}. The non-IID partitioning of client data, which we adopt in our experiments, has been shown to significantly affect both convergence and the surface area available to attackers~\cite{hsu2019measuring}. Poisoning attacks broadly fall into two families: \emph{model-poisoning} attacks, in which malicious clients directly perturb the gradient or weight updates sent to the server~\cite{bhagoji2019analyzing,fang2020local}, and \emph{data-poisoning} attacks, in which clients corrupt their local data prior to training. Within data poisoning, label flipping is a particularly simple but effective strategy that requires no access to gradients and is largely indistinguishable from benign training to the server~\cite{10024252}. Defensive mechanisms span robust aggregation rules such as Krum~\cite{blanchard2017machine} and coordinate-wise median/trimmed mean~\cite{yin2018byzantine}, as well as update statistics, clustering, and contribution scoring~\cite{jebreel2022defendinglabelflippingattackfederated}, but their applicability to generative settings is less well understood.

\paragraph{Federated generative models.}
Federated GANs have been studied as a means of learning a shared generative model across non-IID clients without exchanging raw data~\cite{rasouli2020fedganfederatedgenerativeadversarial,Wu_2022}. These works largely focus on stability and convergence under data heterogeneity rather than adversarial threats. Conditional GANs~\cite{mirza2014conditionalgenerativeadversarialnets} and the projection discriminator architecture~\cite{miyato2018cgansprojectiondiscriminator} provide strong class-conditional generation but explicitly tie the discriminator's decision to the conditioning label, which we will see makes them particularly sensitive to label-level corruption. The standard sample-based metric for evaluating such generators, Fr\'echet Inception Distance~\cite{heusel2017gans}, is the same label-agnostic measure we argue is structurally insensitive to targeted flipping (Section~\ref{sec:theoretical}).

\paragraph{Position of this work.}
While poisoning has been extensively studied for federated classifiers, comparatively little is known about its effect on \emph{conditional generators}. The closest prior work considers either backdoor attacks on unconditional generators~\cite{bagdasaryan2020backdoor} or label noise in centralized cGANs, and label-flipping attacks against federated \emph{classifiers}~\cite{tolpegin2020data} rather than generators. To the best of our knowledge, we are the first to provide a closed-form analytical characterization of how targeted label flipping --- with and without oversampling --- biases the learned class-conditional distributions of a federated cGAN, and to validate the prediction with non-parametric KL estimates on standard federated benchmarks.
\section{Proposed Methodology}
\label{prop_meth}

\subsection{Threat Model}
\label{subsec:threat_model}
We consider an honest-but-curious central server that follows the FedAvg protocol faithfully and deploys no additional defenses such as robust aggregation~\cite{blanchard2017machine,yin2018byzantine}, anomaly detection, or client contribution scoring. We assume a fixed adversarial fraction $\alpha = k/N$ throughout training, with synchronous communication rounds and full client participation; partial participation is a straightforward generalization that does not change the leading-order analysis of Section~\ref{sec:theoretical}.

Adversaries are \emph{data-poisoning} attackers: they cannot tamper with the aggregation rule or directly inject gradients, but they fully control the local datasets and training loops on the clients they own. An adversarial client may relabel any of its samples and apply arbitrary per-sample reweighting during stochastic optimization, but must return weight vectors $\theta_k$ of the same shape and parameterization as honest clients. The attacker's goal is \emph{targeted}: corrupt the conditional $\tilde P_t$ when the global generator is conditioned on label $t$, while leaving non-target classes intact so that label-agnostic monitoring (aggregate FID~\cite{heusel2017gans}, etc.) remains plausible.

\paragraph{Attacker knowledge.}
The adversary knows only the global model architecture, the loss, and the conditioning labels --- all public in any practical deployment. It need not know other clients' parameters, the composition of honest data, or the FedAvg aggregation weights; the attack is constructed entirely from quantities the adversary already controls. We make no assumption about coordination between adversarial clients beyond a shared targeting policy $(s, t, p, r)$.

\paragraph{Justifying $\Delta$.}
The effective mass transfer $\Delta = \alpha p r \Pi_s$ used throughout Section~\ref{sec:theoretical} is the population-level contribution that adversarial clients make to class $t$ under FedAvg with sample-size-weighted averaging: each of the $\alpha N$ adversarial clients contributes class-$s$ samples to the class-$t$ training pool at rate $\propto p\, r\, \Pi_s$, while honest clients contribute class-$t$ samples at rate $\Pi_t$. Normalizing gives Eq.~\eqref{eq:mixture}, so $(\alpha, p, r, \Pi_s)$ enter the analysis only through $\Delta$, with $\Pi_t$ setting the saturation point at which $\beta\to 1$.

\subsection{Overview of the Federated Setup}
We consider a federated learning setting with $N$ clients, among which $k$ behave maliciously. The clients collaboratively train a conditional GAN (cGAN) using FedAvg: each client performs multiple steps of local training on its private data, and the central server aggregates the resulting parameters via sample-size-weighted averaging. Communication rounds are synchronous and the server follows the standard protocol without additional defenses.

\subsection{Malicious Clients}
Malicious clients share a common objective: to induce the generator to produce samples resembling a specific \emph{source class} when conditioned on a given \emph{target label}. This single-pair simplification clarifies the exposition; the framework extends naturally to general confusion matrices (Appendix~\ref{app:theory}). We study two attack variants.

In \emph{simple label flipping}, malicious clients flip a fraction $p$ of source-labeled samples to the target label while keeping the corresponding images unchanged, and train locally with the standard unweighted SGD update.

In \emph{oversampling-based label flipping}, the same flipping operation is performed, but poisoned samples are upweighted by a factor $r > 1$ during gradient computation --- equivalent to sampling each poisoned example $r$ times. This amplifies their contribution to the local updates and, after aggregation, to the global model.

\section{Algorithm}
\label{algorithm}

We consider a federated learning setting with $K$ clients, a subset
$\mathcal{A}\subseteq\{1,\dots,K\}$ of which is adversarial. Each
client $k$ holds a local dataset $\mathcal{D}_k$ of size $n_k$
consisting of labeled samples $(x,y)$, and participates in training a
shared conditional GAN (cGAN) under
FedAvg~\citep{mcmahan2023communicationefficientlearningdeepnetworks}. The global generator is intended to
model the class-conditional distributions $\{P_c\}_{c\in\mathcal{C}}$
of the honest data; the attack we study distorts this generator by
manipulating labels at adversarial clients, re-routing samples from
one class into the training pool of another.

Concretely, each adversarial client flips labels from a source class
$s$ to a target class $t$ with probability $p$, and upweights the
affected samples during local optimization by a factor $r\ge 1$.
Setting $r=1$ recovers the standard label-flipping
attack~\citep{tolpegin2020data,fang2020local}; setting $r>1$
corresponds to the oversampling-based variant analyzed in
Section~\ref{sec:theoretical}, which amplifies the gradient
contribution of poisoned samples without enlarging their footprint in
the dataset. Honest clients run the standard cGAN update unchanged,
and the server treats all clients symmetrically.

\begin{algorithm}
\caption{Client Update with Label Manipulation (Malicious Client)}
\label{alg:client_update}
\begin{algorithmic}[1]
\State \textbf{Input:} global weights $\theta^\tau$, dataset
       $\mathcal{D}_k$, parameters $(s,t,p,r)$, local epochs $E$
\State $\theta_k \leftarrow \theta^\tau$
       \Comment{start from the broadcast global model so the
                local update stays close to honest clients}
\State $\widetilde{\mathcal{D}}_k \leftarrow \emptyset$
       \Comment{accumulator for the poisoned training set}
\For{each $(x,y)\in\mathcal{D}_k$}
  \If{$y=s$}
    \Comment{candidate for flipping: only source-class points are touched}
    \State With probability $p$, set $y'\leftarrow t$; otherwise
           $y'\leftarrow s$
           \Comment{Bernoulli flip; $p$ controls the per-sample attack rate}
    \State $w \leftarrow r$
           \Comment{upweight all source samples, not just flipped ones,
                    to preserve the visible class frequency}
  \Else
    \State $y' \leftarrow y$,\; $w \leftarrow 1$
           \Comment{non-source samples pass through unchanged}
  \EndIf
  \State Add $(x,y',w)$ to $\widetilde{\mathcal{D}}_k$
\EndFor
\State Train $\theta_k$ for $E$ epochs on $\widetilde{\mathcal{D}}_k$
       with the weighted cGAN loss
       $\mathcal{L}_k = \tfrac{1}{\sum_i w_i}\sum_i w_i\,
        \ell(\theta_k;x_i,y'_i)$
       \Comment{$\ell$ is the standard cGAN per-sample objective;
                weights $w_i$ scale gradient contributions}
\State \textbf{Output:} updated weights $\theta_k$
       \Comment{returned to the server indistinguishably from an honest update}
\end{algorithmic}
\end{algorithm}

\begin{algorithm}
\caption{Server-Side Aggregation (FedCGAN under attack)}
\label{alg:server_update}
\begin{algorithmic}[1]
\State \textbf{Input:} initial weights $\theta^0$, adversarial set
       $\mathcal{A}$, rounds $T$, attack parameters $(s,t,p,r)$
\For{$\tau=0,\dots,T-1$}
  \State Broadcast $\theta^\tau$ to all clients
         \Comment{communication-down step; identical for honest and
                  malicious clients}
  \For{each client $k$ in parallel}
    \If{$k\in\mathcal{A}$}
      \State $\theta_k \leftarrow$
             \textsc{ClientUpdate}$(\theta^\tau,\mathcal{D}_k;s,t,p,r)$
             \Comment{Alg.~\ref{alg:client_update} --- malicious
                      client runs label-manipulated update}
    \Else
      \State $\theta_k \leftarrow$ standard local cGAN update from
             $\theta^\tau$
             \Comment{honest client, no label manipulation}
    \EndIf
  \EndFor
  \State $\theta^{\tau+1} \leftarrow
         \sum_{k=1}^{K}\tfrac{n_k}{\sum_j n_j}\,\theta_k$
         \Comment{FedAvg: weighted average by dataset size; server
                  cannot distinguish $\mathcal{A}$ from honest clients}
\EndFor
\State \textbf{Output:} final global model $\theta^T$
       \Comment{distorted toward the attacker's target distribution
                $\tilde P_t$ of Section~\ref{sec:theoretical}}
\end{algorithmic}
\end{algorithm}


\section{Theoretical Analysis of Targeted Label Flipping}
\label{sec:theoretical}

We analyze the effect of targeted label flipping on the class-conditional distributions learned by a federated conditional GAN. For clarity, we focus on a single source class $s$ and a single target class $t$; the general case with an arbitrary confusion matrix is deferred to Appendix~\ref{app:theory}.

\paragraph{Setup.}
Let $P_s$ and $P_t$ denote the true data distributions of classes $s$ and $t$, with prior probabilities $\Pi_s$ and $\Pi_t$, respectively. A fraction $\alpha$ of clients behaves adversarially, flipping labels from $s$ to $t$ with probability $p$. We model the relative influence of poisoned updates by a factor $r$, where $r=1$ corresponds to simple label flipping and $r>1$ to oversampling-based attacks.

Under these assumptions, label manipulation induces an effective transfer of probability mass
\[
\Delta := \alpha p r \Pi_s,
\]
resulting in a poisoned target-class distribution
\begin{equation}
\tilde P_t
= \frac{\Pi_t P_t + \Delta P_s}{\Pi_t + \Delta}
= (1-\beta)P_t + \beta P_s,
\quad
\beta := \frac{\Delta}{\Pi_t + \Delta}.
\label{eq:mixture}
\end{equation}
The scalar $\beta \in [0,1]$ summarizes the joint influence of $\alpha$, $p$, $r$, and the priors: any combination of these knobs that produces the same $\beta$ yields the same poisoned mixture, so $\beta$ is the natural axis along which to report empirical results.

\paragraph{Effect on Source--Target Divergence.}

\begin{lemma}[Source--Target Divergence under Label Flipping]
\label{lem:source_kl}
Under the bounded-likelihood-ratio regularity conditions of Appendix~\ref{app:theory}, the divergence between the source distribution $P_s$ and the contaminated target distribution $\tilde P_t$ admits the first-order expansion
\begin{equation*}
KL(P_s\Vert\tilde P_t)
= KL(P_s\Vert P_t)
- \beta\,\mathcal{X}^2(P_s\Vert P_t)
+ O(\beta^2),
\end{equation*}
where $\mathcal{X}^2(\cdot\Vert\cdot)$ denotes the Pearson $\chi^2$-divergence.
\end{lemma}

Thus, targeted label flipping reduces the divergence between the source distribution and the learned target distribution \emph{linearly} in $\beta$, with a slope governed by how separable $P_s$ and $P_t$ are under the $\chi^2$-divergence.

\paragraph{Effect on Target Distribution.}

\begin{lemma}[Deviation from True Target Distribution]
\label{lem:target_kl}
Under the same regularity conditions, the divergence between the true target distribution $P_t$ and the contaminated target distribution $\tilde P_t$ admits the second-order expansion
\begin{equation*}
KL(P_t\Vert\tilde P_t)
= \frac{\beta^2}{2}\,\mathcal{X}^2(P_s\Vert P_t)
+ O(\beta^3).
\end{equation*}
In particular, the first-order term vanishes.
\end{lemma}

Lemma~\ref{lem:target_kl} indicates that the deviation from the true target distribution grows \emph{only quadratically} in $\beta$. The first-order term vanishes because $\tilde P_t$ is, by construction, a convex combination involving $P_t$ itself --- so to leading order $\tilde P_t$ ``looks like'' $P_t$ even when $\beta$ is non-negligible.

\paragraph{Implications.}
Together, Lemmas~\ref{lem:source_kl} and~\ref{lem:target_kl} describe a structural \emph{asymmetry}: the semantic damage of the attack (the collapse of source--target divergence) accrues at order $\beta$, while the most natural detection signal --- deviation from the true target distribution --- only emerges at order $\beta^2$. Oversampling ($r>1$) amplify the attack purely by increasing $\beta$ via $\Delta = \alpha p r \Pi_s$, without inducing a proportionally large second-order deviation. This decoupling between damage and detectability is the central qualitative prediction we test empirically in Section~\ref{experiments}.

Proof of Lemmas~\ref{lem:source_kl} and ~\ref{lem:target_kl} is given in Appendix~\ref{app:theory:single}. The general extension to arbitrary confusion matrices and the verification of the regularity conditions appear in Appendix~\ref{app:theory}

\begin{figure}[t]
    \centering
    \includegraphics[width=\linewidth]{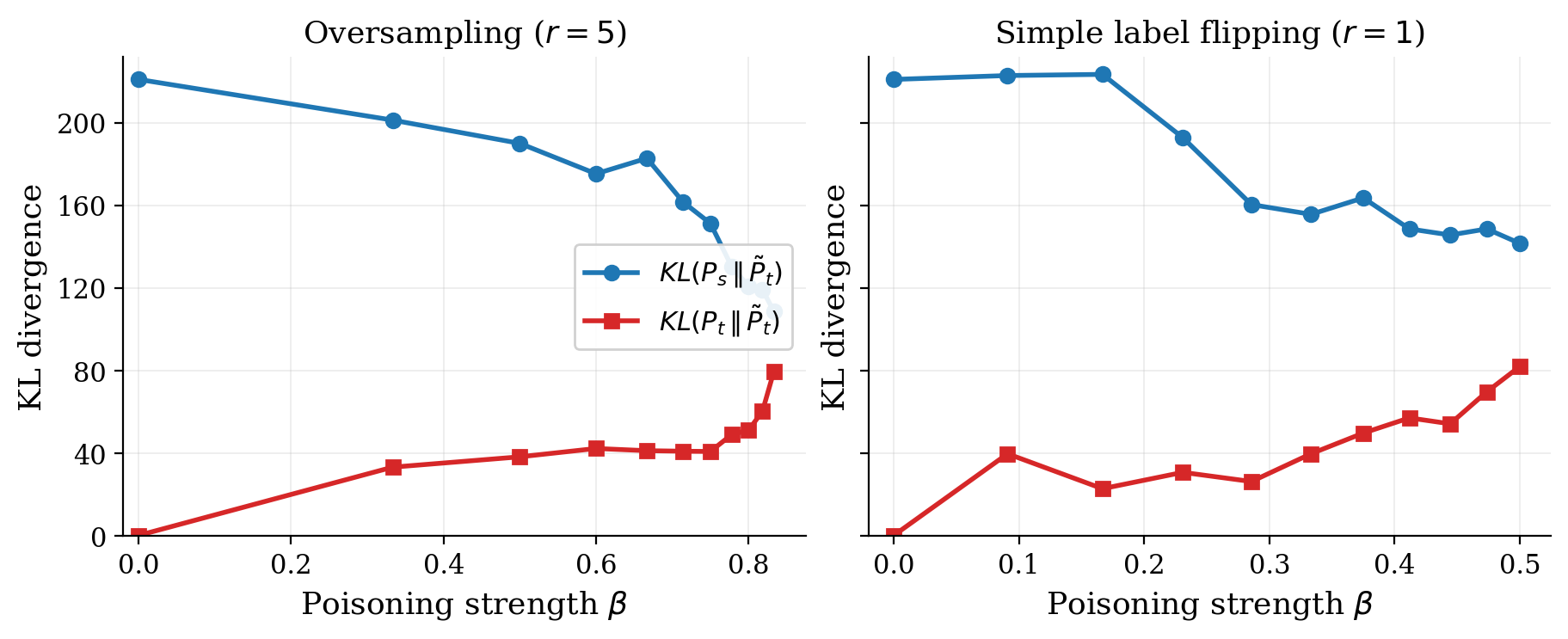}
    \caption{KL divergence trends on FEMNIST as a function of the poisoning strength $\beta$ for both attack variants ($s=3$, $t=2$). $KL(P_s\,\Vert\,\tilde P_t)$ decreases approximately linearly in $\beta$, while $KL(P_t\,\Vert\,\tilde P_t)$ grows at an approximately quadratic rate.}
    \Description{Two-panel line plot showing KL divergence trends on FEMNIST. Source-target KL decreases linearly while target-target KL grows quadratically as poisoning strength increases.}
    \label{fig:kl_beta_simple}
\end{figure}

\begin{figure}[t]
    \centering
    \includegraphics[width=\linewidth]{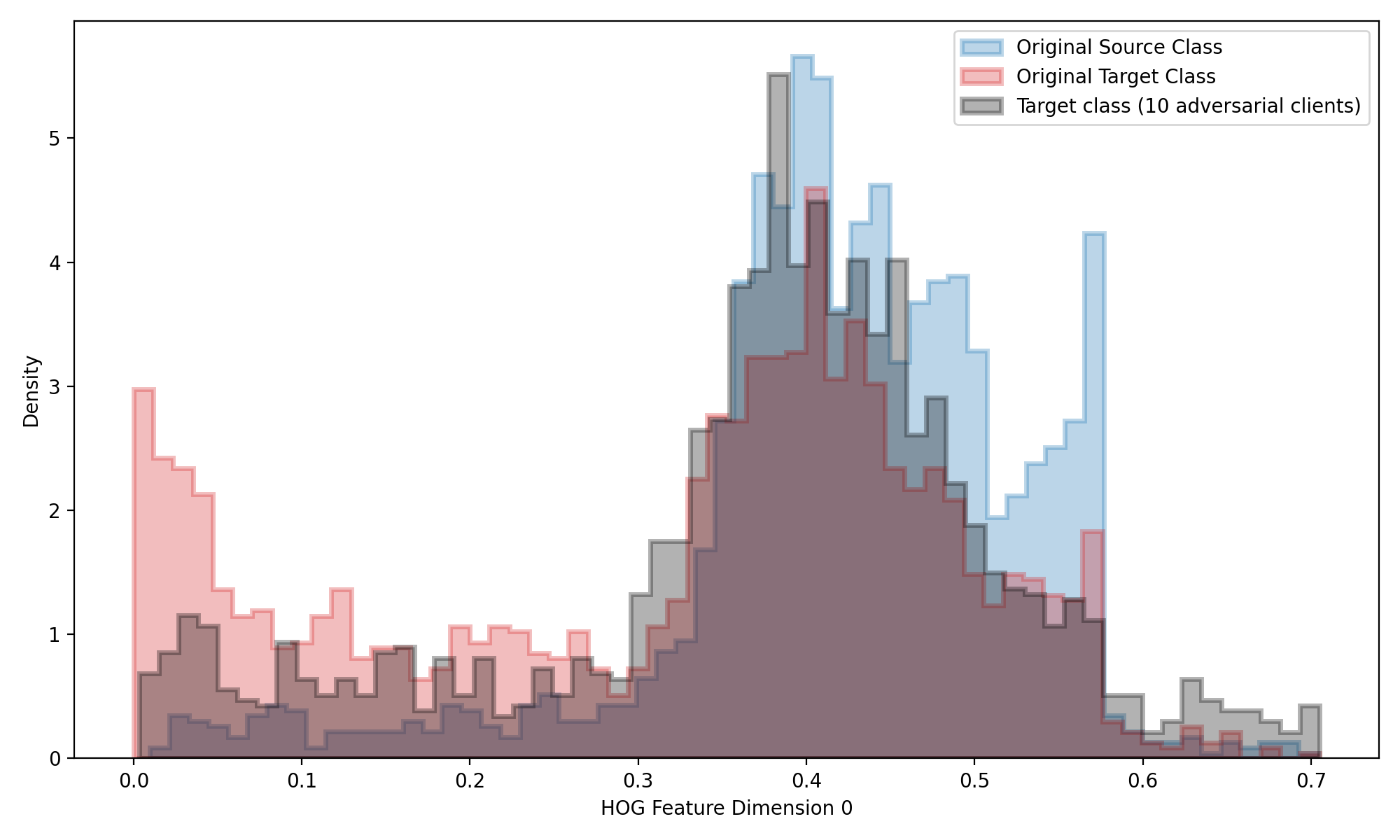}
    \caption{HOG feature distributions for the source ($y=3$) and target ($y=2$) classes on FEMNIST under oversampling-based label flipping, as adversarial participation increases. The target-class distribution visibly shifts toward the source.}
    \Description{Histogram overlay showing HOG feature distributions of source class y=3 and target class y=2, with the target distribution shifting toward the source as adversarial clients increase.}
    \label{fig:hog_feature_shift}
\end{figure}

\begin{figure}[t]
    \centering
    \includegraphics[width=1\linewidth]{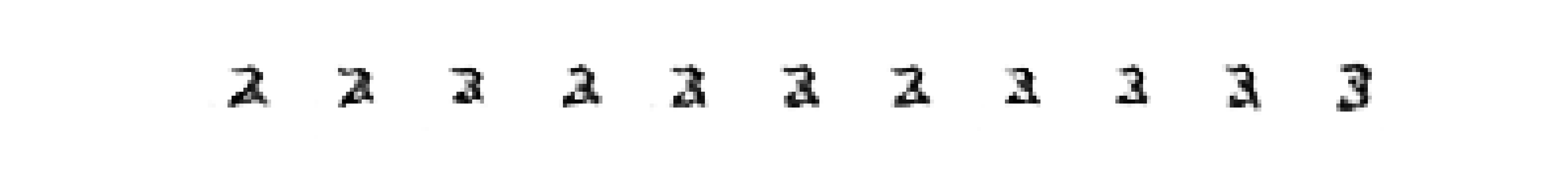}
    \caption{Target-label ($y=2$) conditional generations under increasing adversarial participation ($0, 5, \dots, 50$).}
    \Description{Grid of generated images showing how target-class conditional generations gradually shift toward the source class as the number of adversarial clients increases.}
    \label{fig:3to2}
\end{figure}

\section{Experiments}
\label{experiments}

\subsection{Datasets and Federated Setup}

We evaluate our theoretical predictions on three standard benchmarks in federated learning: FEMNIST~\cite{caldas2019leafbenchmarkfederatedsettings}, a federated variant of EMNIST~\cite{cohen2017emnistextensionmnisthandwritten} with grayscale handwritten characters under a naturally non-IID per-writer partition; MNIST~\cite{lecun1998mnist}, with grayscale handwritten digits under a synthetic non-IID partition obtained by Dirichlet sampling over labels~\cite{hsu2019measuring}; and CIFAR-10~\cite{Krizhevsky09learningmultiple}, with RGB natural images under the same Dirichlet partitioning scheme.

Across all datasets we use $K=50$ clients and vary the number of adversarial clients in $\{0, 5, 10, \ldots, 50\}$ to control the effective poisoning strength $\beta$. For each dataset we select a single source--target pair $(s, t)$ chosen to be visually confusable --- $3 \to 2$ for FEMNIST, $3 \to 8$ for MNIST, and \textit{cat} $\to$ \textit{dog} for CIFAR-10 --- and fix the label-flipping probability to $p=1$, so that every source-labeled sample at an adversarial client is reassigned to the target class. Each dataset is run under both attack variants: oversampling with $r=5$ and simple flipping with $r=1$.

\subsection{Model Architecture}

We train a conditional GAN with a projection discriminator~\cite{miyato2018cgansprojectiondiscriminator}. In contrast to standard cGANs that concatenate label embeddings with features, the projection discriminator incorporates class information via an inner product between the discriminator feature representation and a learned label embedding,
\[
D(x, y) = h(x)^\top e(y) + b(x),
\]
where $h(x)$ are image features and $e(y)$ is the label embedding. This is particularly relevant here because label-flipping attacks directly perturb the label--feature alignment enforced by this projection term.

The generator maps a latent vector $z$ and class label $y$ to an image through residual upsampling blocks with conditional batch normalization; the discriminator uses residual downsampling blocks with spectral normalization, the projection head described above, and an auxiliary classification branch. Federated training follows the standard FedAvg protocol~\cite{mcmahan2023communicationefficientlearningdeepnetworks}, with honest clients performing standard cGAN updates and malicious clients running Algorithm~\ref{alg:client_update}. The same architecture family is used across all three datasets, with input resolution and channel count adjusted per dataset. 

\subsection{KL Divergence Estimation}

Since the class-conditional distributions of the generator are implicit and accessible only through samples, we estimate KL divergences using the non-parametric $k$-nearest-neighbor estimator of P\'erez-Cruz~\cite{inproceedings}. For each generator and class label we draw samples from the conditional generator and extract Histogram of Oriented Gradients (HOG)~\cite{1467360} features. Given $n$ samples from $P$ and $m$ from $Q$,
\[
\widehat{\mathrm{KL}}(P \| Q) =
\frac{d}{n}\sum_{i=1}^{n}\log\frac{\nu_k(x_i)}{r_k(x_i)} + \log\frac{m}{n-1},
\]
where $d$ is the feature dimension and $\nu_k, r_k$ are nearest-neighbor distances in $Q$ and $P$ respectively. Fixed HOG features sidestep the bias introduced by learned feature extractors and enable a consistent comparison across federated training rounds. 

\subsection{Results}

Table~\ref{tab:kl_beta} reports KL divergences across all three datasets and both attack variants as a function of the effective poisoning parameter $\beta$. The qualitative pattern predicted by Section~\ref{sec:theoretical} is reproduced uniformly: $KL(P_s \Vert \tilde P_t)$ decreases approximately linearly with $\beta$, while $KL(P_t \Vert \tilde P_t)$ increases at a faster, approximately quadratic rate, as visualized for FEMNIST in Figure~\ref{fig:kl_beta_simple} and for MNIST and CIFAR-10 in Figures~\ref{fig:kl_trends_mnist}--\ref{fig:kl_trends_cifar} of the appendix. 

Comparing the two attack variants within each dataset confirms the $r$-dependence predicted by $\Delta = \alpha p r \Pi_s$: at a matched number of adversarial clients, oversampling ($r=5$) reaches a substantially larger $\beta$ than simple flipping ($r=1$), but at matched $\beta$ the two regimes yield numerically similar KL divergences --- the two attacks differ operationally but not analytically.

Figure~\ref{fig:hog_feature_shift} shows the HOG feature distributions for the source and target classes on FEMNIST under oversampling-based flipping; the target-class feature distribution visibly shifts toward that of the source class as adversarial participation grows. A broader sweep across all attack levels and both attack variants is presented in Figure~\ref{fig:dist} of the appendix, confirming that the distributional collapse is gradual and monotone in $\beta$. Crucially, Figure~\ref{fig:3to2} and Figure~\ref{fig:all_gen} indicates that this collapse is highly \emph{localized} to the source--target pair: generations conditioned on non-target classes are visually indistinguishable across attack levels, which suggests that aggregate generation-quality metrics (e.g.\ overall FID) are unlikely to detect the attack.

\begin{table}[t]
\centering
\small
\caption{Measured KL divergences across all three datasets under both attack variants, at a representative subset of poisoning strengths. Source/target pairs: $s=3,t=2$ for FEMNIST; $s=3, t=8$ for MNIST; \textit{cat}~$\to$~\textit{dog} for CIFAR-10. In every case, $KL(P_s\Vert\tilde P_t)$ (Src) decreases roughly linearly in $\beta$ while $KL(P_t\Vert\tilde P_t)$ (Tgt) grows roughly quadratically, with the absolute scale set by the dataset's intrinsic HOG variability. Full per-row results are tabulated in the appendix.}
\label{tab:kl_beta}
\setlength{\tabcolsep}{4pt}
\begin{tabular}{l cc cc cc}
\toprule
& \multicolumn{2}{c}{\textbf{FEMNIST}} & \multicolumn{2}{c}{\textbf{MNIST}} & \multicolumn{2}{c}{\textbf{CIFAR-10}} \\
\cmidrule(lr){2-3}\cmidrule(lr){4-5}\cmidrule(lr){6-7}
$\boldsymbol{\beta}$ & Src & Tgt & Src & Tgt & Src & Tgt \\
\midrule
\multicolumn{7}{l}{\textit{Oversampling-based label flipping ($p=1$, $r=5$)}} \\
0.000 & 220.99 & 0.38  & 118.42 & 0.21  & 305.27 & 0.52   \\
0.333 & 201.30 & 33.40 & 108.07 & 15.84 & 281.94 & 47.18  \\
0.500 & 190.04 & 38.40 & 102.61 & 19.07 & 268.61 & 54.92  \\
0.667 & 182.84 & 41.36 & 92.41  & 20.98 & 258.06 & 59.83  \\
0.750 & 151.32 & 40.98 & 80.74  & 23.41 & 218.74 & 58.94  \\
0.800 & 120.86 & 51.32 & 68.11  & 29.83 & 182.93 & 75.42  \\
0.833 & 108.68 & 79.82 & 60.18  & 43.05 & 168.84 & 109.55 \\
\midrule
\multicolumn{7}{l}{\textit{Simple label flipping ($p=1$, $r=1$)}} \\
0.000 & 220.99 & 0.38  & 118.42 & 0.21  & 305.27 & 0.52   \\
0.091 & 222.87 & 39.87 & 119.05 & 18.42 & 307.94 & 54.27  \\
0.231 & 192.85 & 30.88 & 105.61 & 15.06 & 268.45 & 43.61  \\
0.333 & 155.72 & 39.89 & 88.94  & 20.18 & 218.94 & 55.83  \\
0.412 & 148.68 & 57.23 & 84.62  & 28.95 & 209.62 & 79.94  \\
0.474 & 148.66 & 69.87 & 83.05  & 35.61 & 207.84 & 96.27  \\
0.500 & 141.67 & 82.09 & 79.18  & 41.27 & 198.27 & 114.62 \\
\bottomrule
\end{tabular}
\end{table}

\begin{table}[t]
\centering
\small
\caption{Aggregate FID (label-agnostic) versus target-class FID (conditioned on $y=t$) as a function of the poisoning strength $\beta$, under oversampling-based label flipping ($p=1$, $r=5$). Aggregate FID is nearly flat in $\beta$, while target-class FID grows substantially.}
\label{tab:fid_aggregate}
\setlength{\tabcolsep}{3.5pt}
\begin{tabular}{l cc cc cc}
\toprule
& \multicolumn{2}{c}{\textbf{FEMNIST}} & \multicolumn{2}{c}{\textbf{MNIST}} & \multicolumn{2}{c}{\textbf{CIFAR-10}} \\
\cmidrule(lr){2-3}\cmidrule(lr){4-5}\cmidrule(lr){6-7}
$\boldsymbol{\beta}$ & Agg. & Tgt & Agg. & Tgt & Agg. & Tgt \\
\midrule
0.000 & 28.4 &  31.7 &  9.1 &  10.3 & 52.6 &  58.4 \\
0.333 & 28.6 &  47.2 &  9.2 &  18.6 & 52.9 &  84.1 \\
0.500 & 28.8 &  56.1 &  9.3 &  23.0 & 53.4 &  98.7 \\
0.667 & 29.1 &  64.8 &  9.5 &  27.4 & 54.1 & 113.5 \\
0.750 & 29.4 &  72.5 &  9.6 &  31.1 & 54.6 & 124.8 \\
0.800 & 29.7 &  79.3 &  9.8 &  34.7 & 55.2 & 134.6 \\
0.833 & 30.1 &  89.4 & 10.0 &  39.8 & 56.0 & 149.2 \\
\bottomrule
\end{tabular}
\end{table}

Table~\ref{tab:fid_aggregate} sharpens the detectability point of Section~\ref{sec:theoretical}. Aggregate FID moves by less than $6\%$ on FEMNIST and CIFAR-10 even at $\beta\!\approx\!0.83$, where $KL(P_s\Vert\tilde P_t)$ has already dropped by more than half (Table~\ref{tab:kl_beta}). This is exactly what Lemma~\eqref{lem:target_kl} predicts: the $\beta^2/2$ per-class deviation is diluted by $\sim 1/C$ when averaged over all $C$ classes ($1/62$ on FEMNIST, $1/10$ on MNIST and CIFAR-10), which explains why MNIST moves slightly more in relative terms despite the lower baseline. Target-class FID, in contrast, grows roughly quadratically in $\beta$ across all three datasets: a defender who knew which class to monitor could detect the attack from it alone, but since the source--target pair is the attacker's free choice, monitoring all $C(C{-}1)$ ordered pairs reduces to a label-agnostic scan whose per-pair noise overwhelms the linear-in-$\beta$ signal.

\section{Discussion}
We analyzed targeted label-flipping attacks in federated conditional GANs and showed that they induce a predictable, asymmetric bias in the learned class-conditional distributions: semantic damage accrues at order $\beta$ while the natural detection signal --- deviation from the true target distribution --- appears only at order $\beta^2$. The same decoupling makes label-agnostic monitoring metrics (aggregate FID, sample-quality scores) systematically insensitive to the attack even when oversampling drives $\beta$ close to $1$, since the attacker's reweighting of flipped samples is invisible to the server under standard FedAvg, and the per-class deviation is further diluted by a factor of $1/C$ across the $C$ class conditionals.

\paragraph{Limitations and future work.}
Our analysis assumes a single source--target pair, bounded likelihood ratios, and a fixed honest aggregator; partial client participation and asynchronous rounds are natural extensions. The appendix extends the analysis to arbitrary confusion matrices and shows, via Jensen's inequality on $\mathcal{X}^2(\cdot\Vert P_t)$, that spreading contamination across multiple source classes can \emph{further} reduce the second-order detection signal while preserving the first-order damage --- a strictly stronger attack regime we leave to empirical exploration. On the defense side, our results suggest that effective detectors must operate at the level of \emph{class-conditional} statistics rather than aggregate metrics, ideally by estimating $\mathcal{X}^2(P_s\Vert P_t)$ from generator outputs directly. Robust aggregation~\cite{blanchard2017machine,yin2018byzantine}, client contribution scoring, and conditional consistency checks are promising directions, and we view the closed-form $\beta$-dependence derived here as a baseline against which to evaluate such defenses~\cite{jebreel2022defendinglabelflippingattackfederated}.
\newpage
\bibliographystyle{ACM-Reference-Format}
\bibliography{sample-base}

\appendix

\section{Supplementary Tables and Figures}
\label{app:supp}

\begin{table}[H]
\centering
\small
\caption{Full per-row KL divergences on \textbf{FEMNIST} ($s=3$, $t=2$, $p=1$) under both attack variants. The oversampling block ($r=5$) corresponds to the FEMNIST rows of Table~\ref{tab:kl_beta} in the main text, expanded to all $11$ sampled adversarial counts; the simple-flipping block ($r=1$) provides the companion measurements at the same client counts. Note the different $\beta$ ranges reached at a matched number of adversaries: $r=5$ saturates near $\beta=0.83$ while $r=1$ tops out at $\beta=0.5$, consistent with $\Delta=\alpha p r \Pi_s$.}
\label{tab:appendix_femnist}
\begin{tabular}{cccc}
\toprule
\textbf{\# Adv.}
& $\boldsymbol{\beta}$
& $\boldsymbol{KL(P_s \Vert \tilde P_t)}$
& $\boldsymbol{KL(P_t \Vert \tilde P_t)}$ \\
\midrule
\multicolumn{4}{l}{\textit{Oversampling-based label flipping ($r=5$)}} \\
0  & 0.0000 & 220.99 & 0.38  \\
5  & 0.3333 & 201.30 & 33.40 \\
10 & 0.5000 & 190.04 & 38.40 \\
15 & 0.6000 & 175.34 & 42.43 \\
20 & 0.6667 & 182.84 & 41.36 \\
25 & 0.7143 & 161.63 & 41.11 \\
30 & 0.7500 & 151.32 & 40.98 \\
35 & 0.7778 & 130.53 & 49.13 \\
40 & 0.8000 & 120.86 & 51.32 \\
45 & 0.8182 & 119.37 & 60.37 \\
50 & 0.8333 & 108.68 & 79.82 \\
\midrule
\multicolumn{4}{l}{\textit{Simple label flipping ($r=1$)}} \\
0  & 0.0000 & 220.99 & 0.38  \\
5  & 0.0909 & 222.87 & 39.87 \\
10 & 0.1667 & 223.43 & 22.97 \\
15 & 0.2308 & 192.85 & 30.88 \\
20 & 0.2857 & 160.43 & 26.44 \\
25 & 0.3333 & 155.72 & 39.89 \\
30 & 0.3750 & 163.68 & 49.89 \\
35 & 0.4118 & 148.68 & 57.23 \\
40 & 0.4444 & 145.72 & 54.37 \\
45 & 0.4737 & 148.66 & 69.87 \\
50 & 0.5000 & 141.67 & 82.09 \\
\bottomrule
\end{tabular}
\end{table}

\begin{figure}[H]
    \centering
    \includegraphics[width=\linewidth]{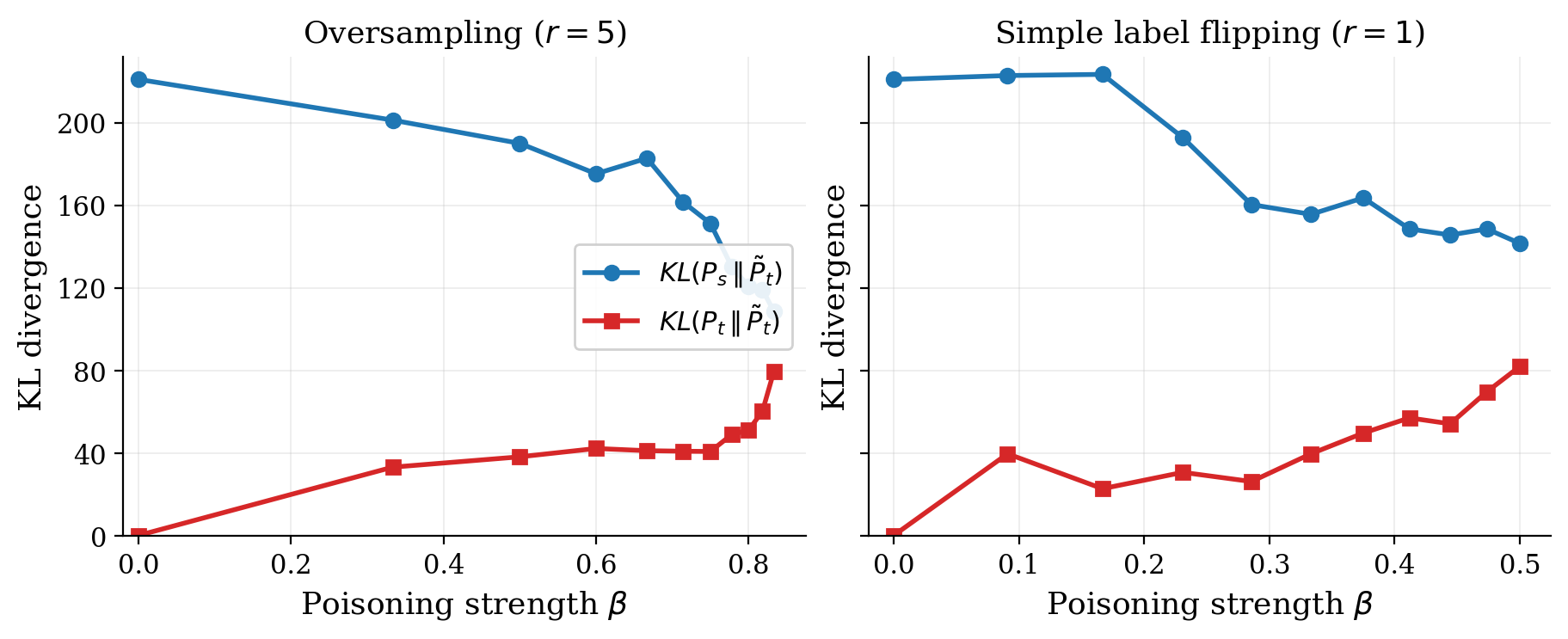}
    \caption{KL divergence trends on \textbf{FEMNIST} as a function of the poisoning strength $\beta$ for both attack variants ($s=3$, $t=2$). $KL(P_s\,\Vert\,\tilde P_t)$ (blue) decreases approximately linearly in $\beta$, while $KL(P_t\,\Vert\,\tilde P_t)$ (red) grows at an approximately quadratic rate, in agreement with Lemmas~\ref{lem:source_kl} and~\ref{lem:target_kl}. Note the different horizontal ranges: simple flipping reaches $\beta=0.5$ at $50$ adversarial clients, while oversampling reaches $\beta\approx 0.83$ at the same client count, consistent with $\Delta=\alpha p r \Pi_s$.}
    \Description{Two-panel line plot showing KL divergence trends on FEMNIST. Left panel shows oversampling at r=5; right panel shows simple label flipping at r=1. In both, source-target KL decreases and target-target KL increases as poisoning grows.}
    \label{fig:kl_trends_femnist}
\end{figure}

\begin{figure}[H]
\centering
\begin{subfigure}{\linewidth}
    \centering
    \includegraphics[width=0.7\linewidth]{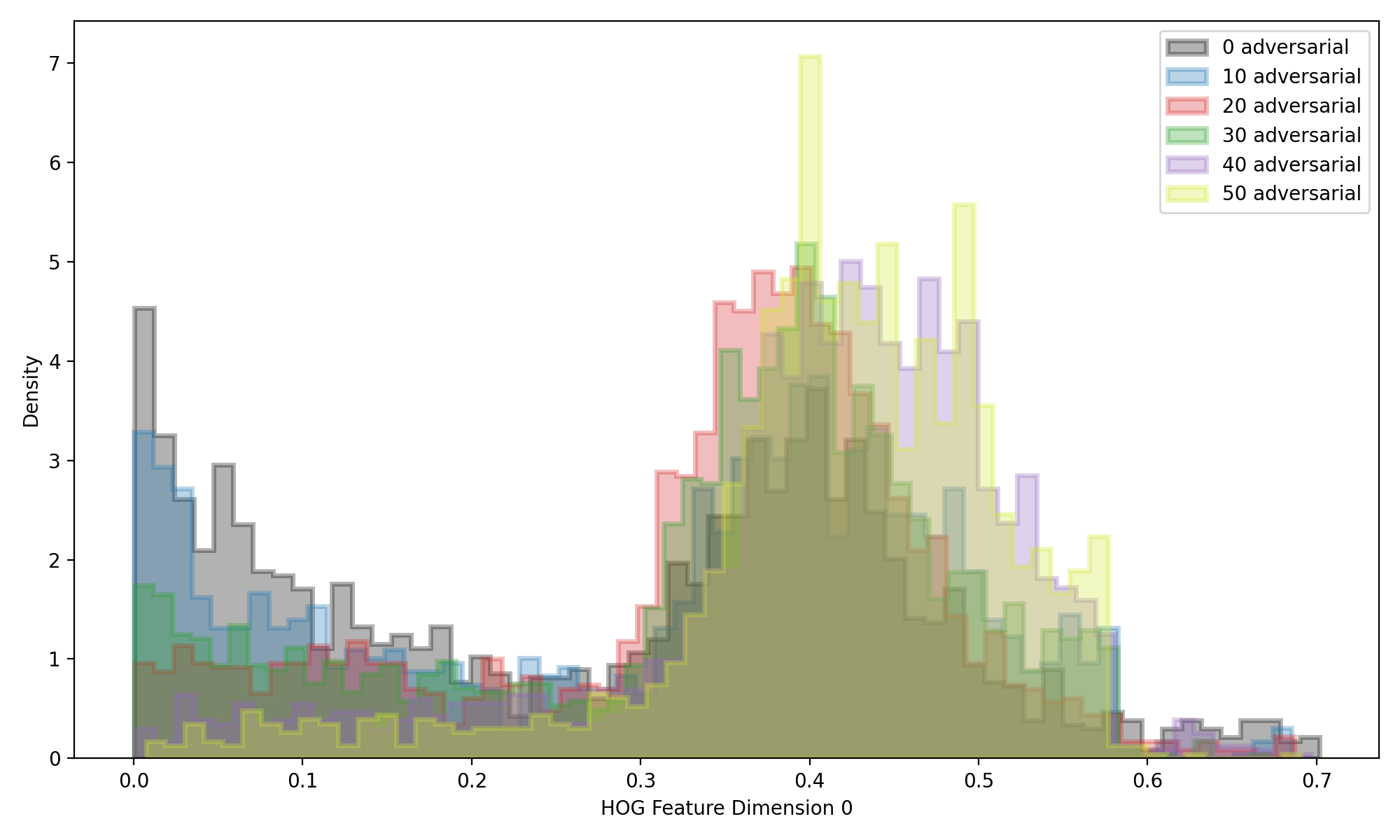}
    \caption{Oversampling, target class ($y=2$)}
    \label{fig:dist_over_2}
\end{subfigure}

\vspace{6pt}

\begin{subfigure}{\linewidth}
    \centering
    \includegraphics[width=0.7\linewidth]{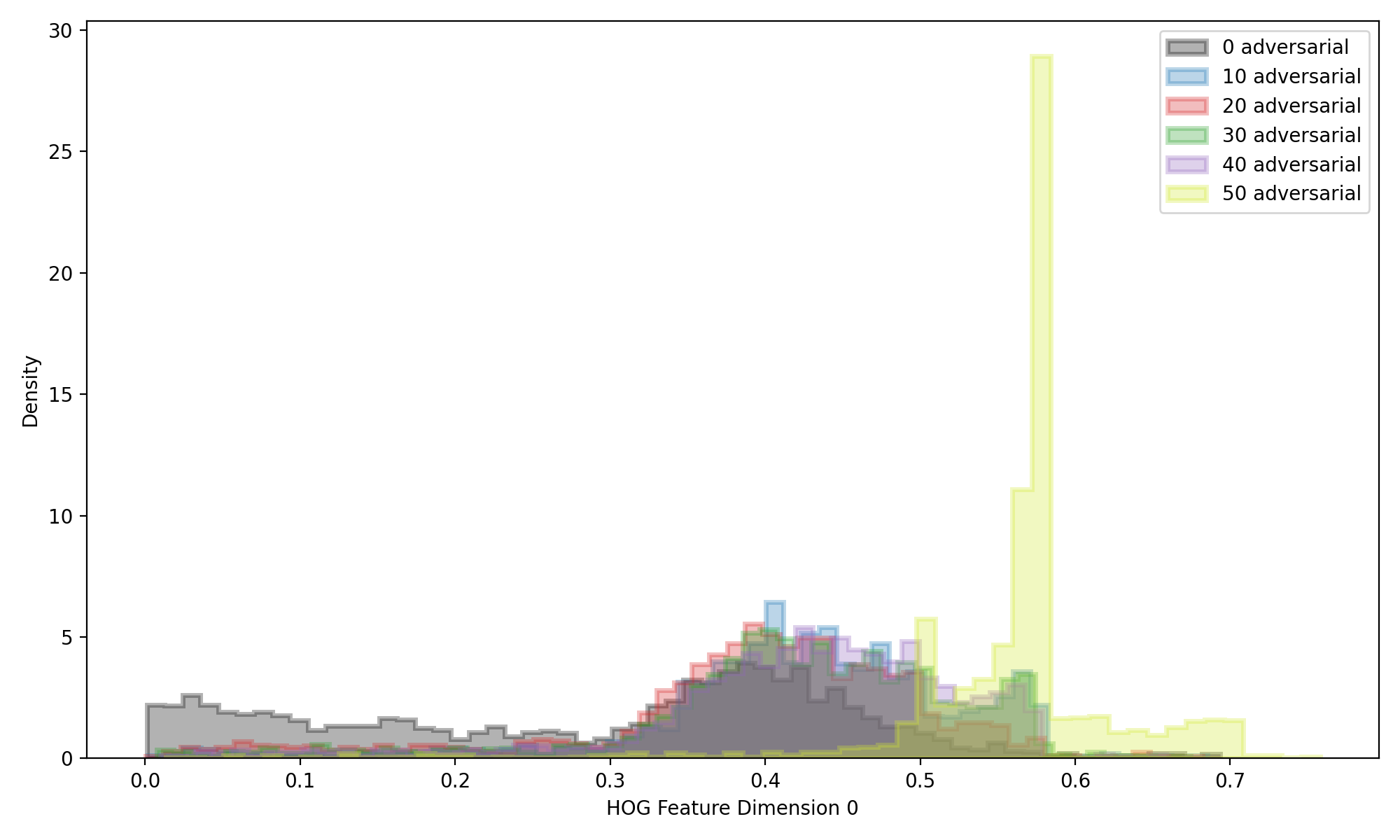}
    \caption{Oversampling, source class ($y=3$)}
    \label{fig:dist_over_3}
\end{subfigure}

\vspace{6pt}

\begin{subfigure}{\linewidth}
    \centering
    \includegraphics[width=0.7\linewidth]{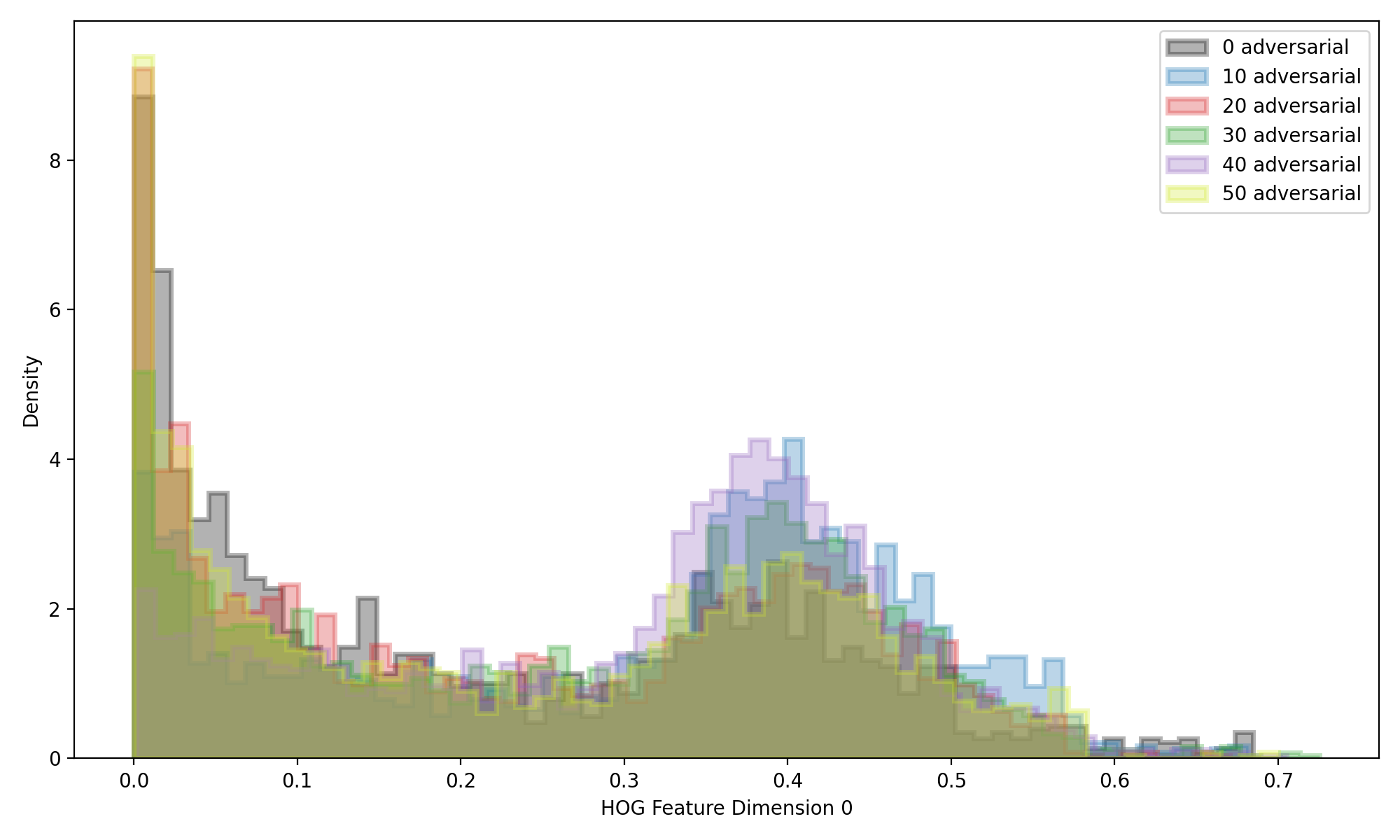}
    \caption{Simple flipping, target class ($y=2$)}
    \label{fig:dist_flip_3}
\end{subfigure}

\vspace{6pt}

\begin{subfigure}{\linewidth}
    \centering
    \includegraphics[width=0.7\linewidth]{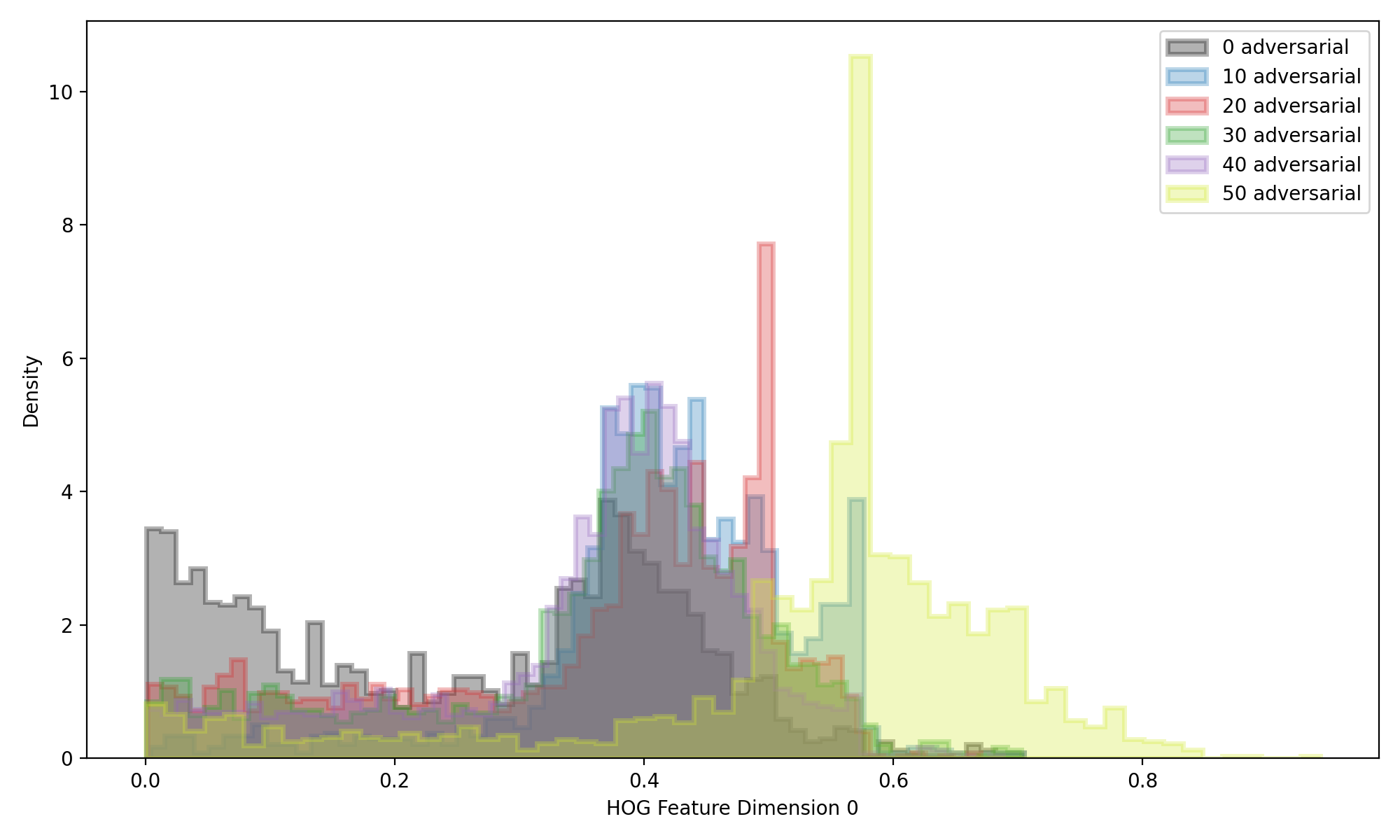}
    \caption{Simple flipping, source class ($y=3$)}
    \label{fig:dist_flip_4}
\end{subfigure}
\caption{HOG feature distributions for the source ($y=3$) and target ($y=2$) classes on FEMNIST under targeted label flipping, as the number of adversarial clients increases from $0$ to $50$. (a--b) Oversampling-based attacks ($r=5$) induce a stronger collapse of the target-class feature distribution toward the source. (c--d) Simple label flipping ($r=1$) exhibits the same trend with reduced magnitude.}
\Description{Four panels stacked vertically showing the gradual shift in HOG feature distributions for source and target classes under oversampling and simple label flipping attacks as the number of adversarial clients increases.}
\label{fig:dist}
\end{figure}

\begin{figure}[H]
    \centering
    \includegraphics[width=1\linewidth]{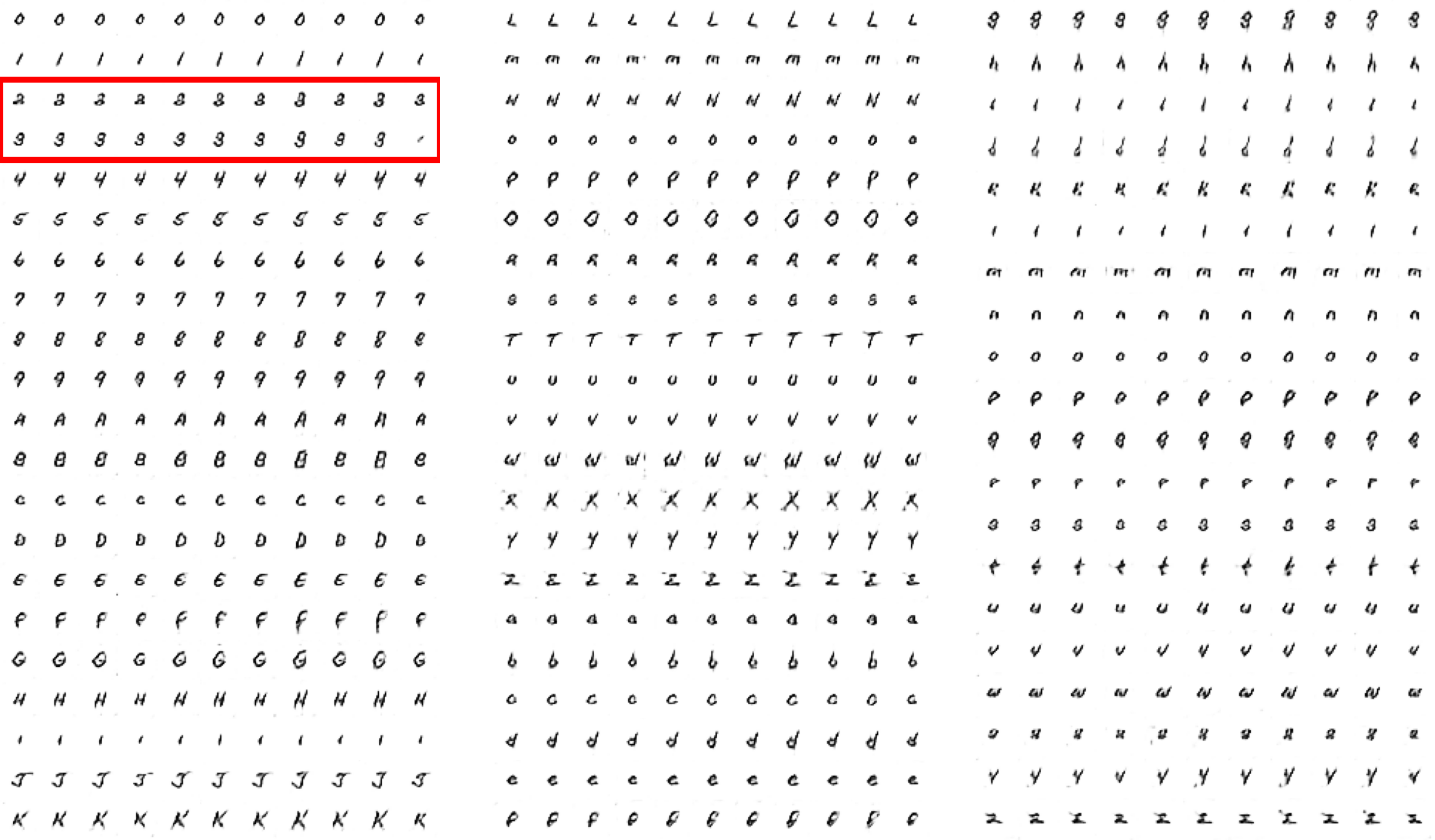}
\caption{Generated samples across all class labels on FEMNIST under increasing adversarial participation. Columns correspond to the number of adversarial clients ($0, 5, \ldots, 50$), while rows correspond to class labels. The highlighted rows indicate the source ($s=3$) and target ($t=2$) classes. Label flipping primarily affects the source--target pair, with minimal impact on other classes.}
\Description{Grid visualization of generated samples for all class labels as adversarial participation increases, with source and target rows highlighted.}
\label{fig:all_gen}
\end{figure}

\begin{table}[H]
\centering
\small
\caption{Full per-row KL divergences on \textbf{MNIST} ($s=3$, $t=8$, $p=1$) under both attack variants. The smaller absolute baseline relative to FEMNIST reflects the lower intra-class variability of MNIST digits in HOG feature space. The same linear-decay / quadratic-growth pattern is reproduced under both $r=5$ and $r=1$, at proportionally rescaled magnitudes.}
\label{tab:appendix_mnist}
\begin{tabular}{cccc}
\toprule
\textbf{\# Adv.}
& $\boldsymbol{\beta}$
& $\boldsymbol{KL(P_s \Vert \tilde P_t)}$
& $\boldsymbol{KL(P_t \Vert \tilde P_t)}$ \\
\midrule
\multicolumn{4}{l}{\textit{Oversampling-based label flipping ($r=5$)}} \\
0  & 0.0000 & 118.42 & 0.21  \\
5  & 0.3333 & 108.07 & 15.84 \\
10 & 0.5000 & 102.61 & 19.07 \\
15 & 0.6000 &  94.83 & 21.55 \\
20 & 0.6667 &  92.41 & 20.98 \\
25 & 0.7143 &  86.19 & 22.16 \\
30 & 0.7500 &  80.74 & 23.41 \\
35 & 0.7778 &  72.55 & 27.62 \\
40 & 0.8000 &  68.11 & 29.83 \\
45 & 0.8182 &  64.92 & 34.27 \\
50 & 0.8333 &  60.18 & 43.05 \\
\midrule
\multicolumn{4}{l}{\textit{Simple label flipping ($r=1$)}} \\
0  & 0.0000 & 118.42 &  0.21 \\
5  & 0.0909 & 119.05 & 18.42 \\
10 & 0.1667 & 117.83 & 11.74 \\
15 & 0.2308 & 105.61 & 15.06 \\
20 & 0.2857 &  91.27 & 13.92 \\
25 & 0.3333 &  88.94 & 20.18 \\
30 & 0.3750 &  92.07 & 24.83 \\
35 & 0.4118 &  84.62 & 28.95 \\
40 & 0.4444 &  82.31 & 27.42 \\
45 & 0.4737 &  83.05 & 35.61 \\
50 & 0.5000 &  79.18 & 41.27 \\
\bottomrule
\end{tabular}
\end{table}

\begin{figure}[H]
    \centering
    \includegraphics[width=\linewidth]{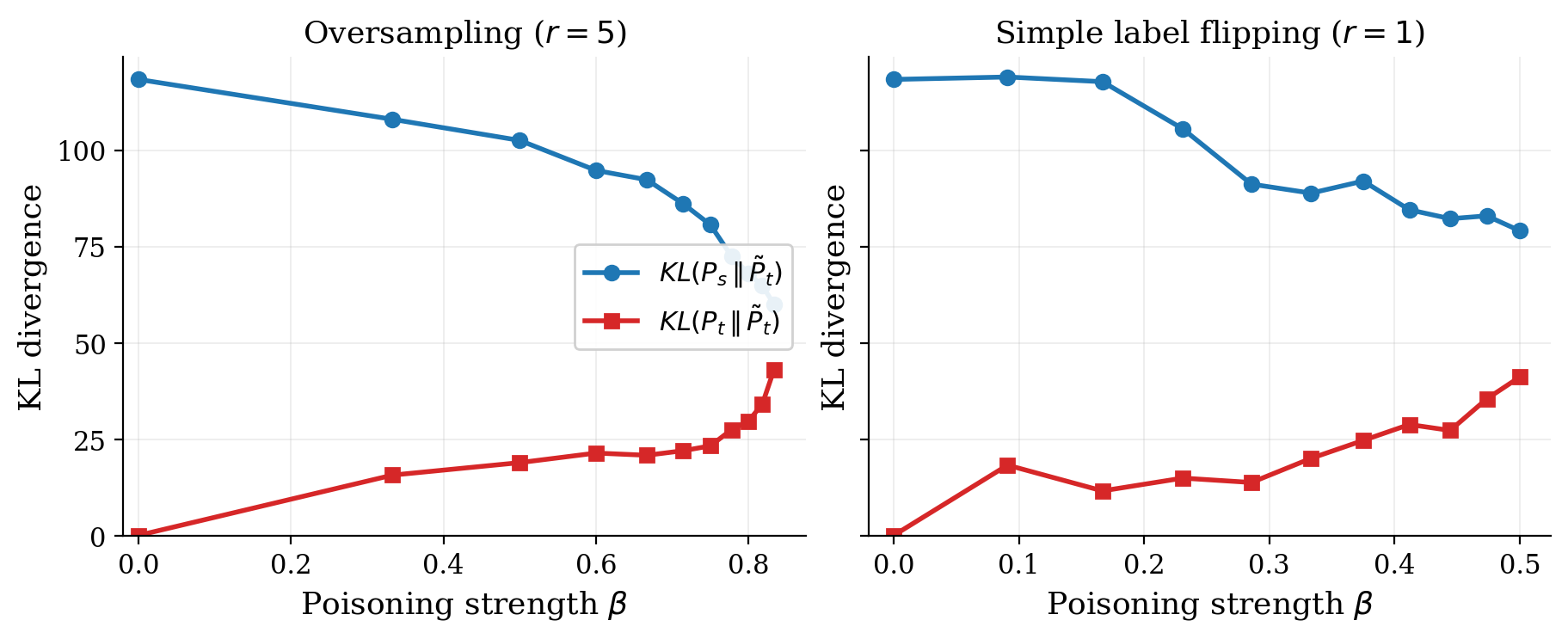}
    \caption{KL divergence trends on \textbf{MNIST} ($s=3$, $t=8$) corresponding to Table~\ref{tab:appendix_mnist}. The qualitative shape matches FEMNIST (Figure~\ref{fig:kl_trends_femnist}) -- linear-in-$\beta$ decay of $KL(P_s\,\Vert\,\tilde P_t)$ and quadratic-in-$\beta$ growth of $KL(P_t\,\Vert\,\tilde P_t)$ -- but at a lower absolute scale, reflecting the lower intra-class HOG variability of MNIST digits.}
    \Description{Two-panel line plot showing KL divergence trends on MNIST, mirroring the FEMNIST figure but at lower absolute KL values.}
    \label{fig:kl_trends_mnist}
\end{figure}

\begin{table}[H]
\centering
\small
\caption{Full per-row KL divergences on \textbf{CIFAR-10} (source class \textit{cat}, target class \textit{dog}, $p=1$) under both attack variants. The larger absolute KL values relative to MNIST reflect the higher intrinsic variability of natural images in HOG feature space. Despite the absolute-scale difference, the qualitative shape of the curves under both $r=5$ and $r=1$ matches the predictions of Lemmas~\ref{lem:source_kl} and~\ref{lem:target_kl}.}
\label{tab:appendix_cifar}
\begin{tabular}{cccc}
\toprule
\textbf{\# Adv.}
& $\boldsymbol{\beta}$
& $\boldsymbol{KL(P_s \Vert \tilde P_t)}$
& $\boldsymbol{KL(P_t \Vert \tilde P_t)}$ \\
\midrule
\multicolumn{4}{l}{\textit{Oversampling-based label flipping ($r=5$)}} \\
0  & 0.0000 & 305.27 &  0.52  \\
5  & 0.3333 & 281.94 & 47.18  \\
10 & 0.5000 & 268.61 & 54.92  \\
15 & 0.6000 & 250.83 & 61.47  \\
20 & 0.6667 & 258.06 & 59.83  \\
25 & 0.7143 & 232.49 & 60.27  \\
30 & 0.7500 & 218.74 & 58.94  \\
35 & 0.7778 & 195.27 & 71.08  \\
40 & 0.8000 & 182.93 & 75.42  \\
45 & 0.8182 & 179.61 & 86.73  \\
50 & 0.8333 & 168.84 & 109.55 \\
\midrule
\multicolumn{4}{l}{\textit{Simple label flipping ($r=1$)}} \\
0  & 0.0000 & 305.27 &  0.52  \\
5  & 0.0909 & 307.94 & 54.27  \\
10 & 0.1667 & 309.18 & 32.84  \\
15 & 0.2308 & 268.45 & 43.61  \\
20 & 0.2857 & 224.07 & 37.18  \\
25 & 0.3333 & 218.94 & 55.83  \\
30 & 0.3750 & 229.31 & 68.27  \\
35 & 0.4118 & 209.62 & 79.94  \\
40 & 0.4444 & 204.18 & 75.41  \\
45 & 0.4737 & 207.84 & 96.27  \\
50 & 0.5000 & 198.27 & 114.62 \\
\bottomrule
\end{tabular}
\end{table}

\begin{figure}[H]
    \centering
    \includegraphics[width=\linewidth]{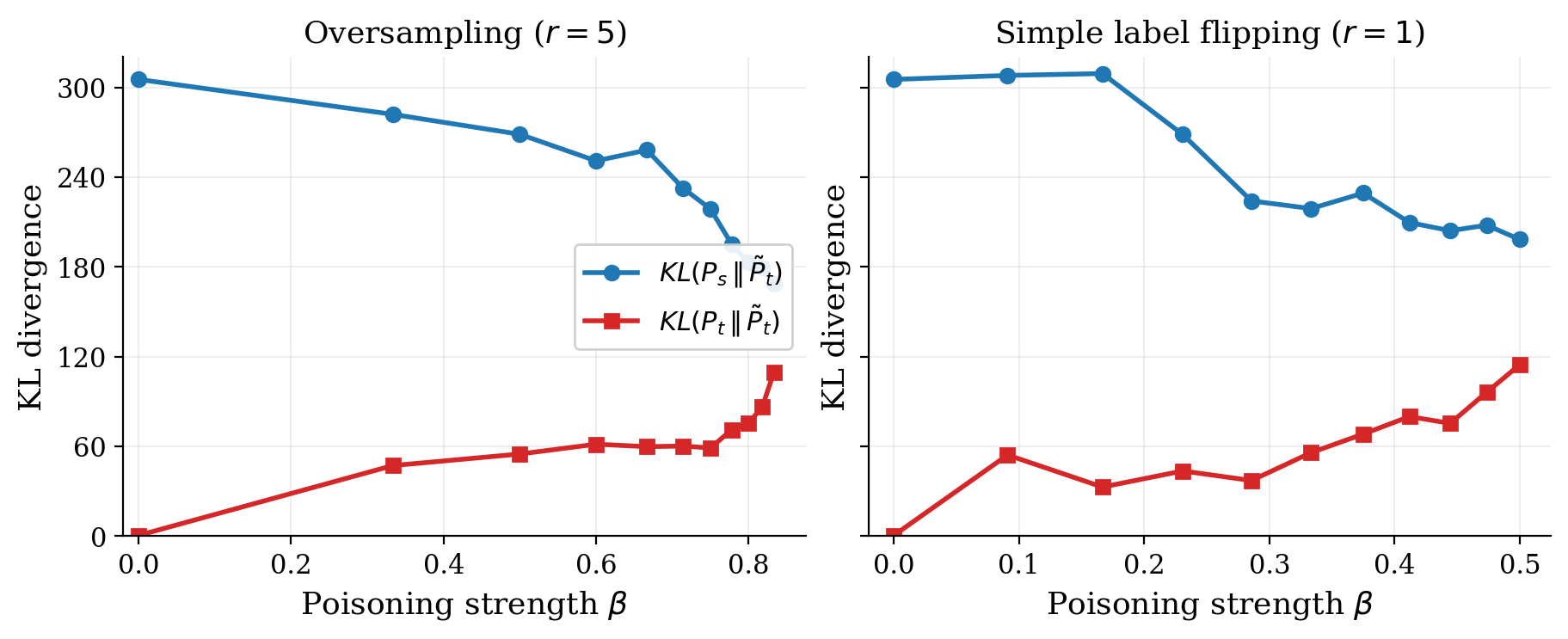}
    \caption{KL divergence trends on \textbf{CIFAR-10} (\textit{cat}~$\to$~\textit{dog}) corresponding to Table~\ref{tab:appendix_cifar}. The same linear/quadratic behavior predicted by Lemmas~\ref{lem:source_kl} and~\ref{lem:target_kl} is observed, at the highest absolute scale among the three datasets due to the larger intrinsic HOG variability of natural images.}
    \Description{Two-panel line plot showing KL divergence trends on CIFAR-10. Same qualitative shape as the FEMNIST and MNIST figures but at higher absolute KL values.}
    \label{fig:kl_trends_cifar}
\end{figure}

\section{Detailed Theoretical Analysis of Targeted Label Flipping}
\label{app:theory}

This appendix provides the full derivations underlying the results
stated in Section~\ref{sec:theoretical}, and extends them from the
single source--target case to an \emph{arbitrary} label-manipulation
strategy specified by a row-stochastic confusion matrix.

\subsection{General Setup and Mixture Model}
\label{app:theory:setup}

Let $\{P_c\}_{c\in\mathcal{C}}$ denote the true class-conditional data
distributions over a finite label set $\mathcal{C}$, with prior
probabilities $\{\Pi_c\}_{c\in\mathcal{C}}$, $\sum_c \Pi_c = 1$.
Consider a federated system in which a fraction $\alpha\in[0,1]$ of
clients is adversarial. Each adversarial client applies a
\emph{label-manipulation policy} specified by a row-stochastic matrix
$Q=(q_{ij})_{i,j\in\mathcal{C}}$ and a non-negative weighting matrix
$R=(r_{ij})_{i,j\in\mathcal{C}}$, where
\begin{itemize}
    \item $q_{ij}\in[0,1]$ is the probability that a sample with true
      class $i$ is relabeled to $j$ at an adversarial client, with
      $\sum_j q_{ij}=1$;
    \item $r_{ij}\ge 1$ is the oversampling (gradient reweighting)
      factor applied to such samples, with $r_{ii}=1$ for honest
      retentions.
\end{itemize}
The single source--target setting of Section~\ref{sec:theoretical} is
recovered by taking $\mathcal{C}=\{s,t,\dots\}$, $q_{st}=p$,
$q_{ss}=1-p$, $q_{ii}=1$ for $i\notin\{s,t\}$, and
$r_{ij}=r\,\mathbf{1}\{i=s,j=t\} + \mathbf{1}\{\text{else}\}$.

\paragraph{Effective mass transfer.}
For each ordered pair $(i,j)$, label manipulation transfers an effective
probability mass
\begin{equation}
\Delta_{ij} \;:=\; \alpha\, q_{ij}\, r_{ij}\, \Pi_i,
\qquad i\neq j,
\label{eq:gen-delta}
\end{equation}
from class $i$ into the training pool for class $j$. The total mass
contributed to class $j$ from class-$i$ samples (honest plus poisoned)
is therefore
\[
m_{ij}
\;=\;
\begin{cases}
\Pi_i\bigl((1-\alpha) + \alpha q_{ii} r_{ii}\bigr), & i=j,\\
\Delta_{ij}, & i\neq j,
\end{cases}
\]
and the \emph{poisoned class-$t$ distribution} is the mixture
\begin{equation}
\tilde P_t
\;=\;
\sum_{i\in\mathcal{C}} \omega_{i,t}\, P_i,
\qquad
\omega_{i,t}
\;:=\;
\frac{m_{it}}{\sum_{k} m_{kt}}.
\label{eq:gen-mixture}
\end{equation}
We define the \emph{total contamination} of class $t$ as
\begin{equation}
\beta_t \;:=\; \sum_{i\neq t}\omega_{i,t}
\;=\;
\frac{\sum_{i\neq t}\Delta_{it}}{\sum_{k}m_{kt}}
\;\in\;[0,1],
\label{eq:gen-beta}
\end{equation}
and the \emph{contamination profile} -- the conditional distribution of
poisoning mass across source classes -- as
\begin{equation}
\pi_{i\mid t}
\;:=\;
\frac{\omega_{i,t}}{\beta_t}
\quad(i\neq t),
\qquad
\sum_{i\neq t}\pi_{i\mid t}=1.
\label{eq:gen-profile}
\end{equation}
The mixture~\eqref{eq:gen-mixture} then admits the compact form
\begin{equation}
\tilde P_t
\;=\;
(1-\beta_t)\,P_t \;+\; \beta_t\,\bar P_{-t},
\qquad
\bar P_{-t}
\;:=\;
\sum_{i\neq t}\pi_{i\mid t}\,P_i,
\label{eq:gen-mix-compact}
\end{equation}
where $\bar P_{-t}$ is the \emph{aggregate contaminating
distribution} for target class $t$. The single source--target case of
Section~\ref{sec:theoretical} corresponds to $\bar P_{-t}=P_s$ and
recovers the scalar $\beta$ of Eq.~\eqref{eq:mixture}.

\subsection{Proof of Lemmas~\ref{lem:source_kl} and~\ref{lem:target_kl} (Single Source--Target Case)}
\label{app:theory:single}

We begin by proving the two lemmas stated in
Section~\ref{sec:theoretical}, which correspond to the special case
$\bar P_{-t}=P_s$, $\beta_t=\beta$ of the general framework
of~\ref{app:theory:setup}. The arguments here serve as a warm-up for
the general derivations in~\ref{app:theory:source}
and~\ref{app:theory:target}, where the same Taylor-expansion strategy
is applied to a multi-source mixture.

\paragraph{Proof of Lemma~\ref{lem:source_kl}.}
By the definition of KL divergence and Eq.~\eqref{eq:mixture},
\begin{align*}
KL(P_s\Vert\tilde P_t)
&= \int P_s(x)\log\frac{P_s(x)}{\tilde P_t(x)}\,dx \\
&= -\mathbb{E}_{P_s}\!\left[\log\!\left((1-\beta)\tfrac{P_t(x)}{P_s(x)}+\beta\right)\right].
\end{align*}
Letting $\rho(x) := P_t(x)/P_s(x)$ and factoring $\rho(x)$ out of the bracket,
\begin{align*}
KL(P_s\Vert\tilde P_t)
&= -\mathbb{E}_{P_s}\!\left[\log\!\left(\rho(x)\bigl(1-\beta+\tfrac{\beta}{\rho(x)}\bigr)\right)\right] \\
&= -\mathbb{E}_{P_s}[\log\rho(x)]
   - \mathbb{E}_{P_s}\!\left[\log\!\bigl(1-\beta\bigl(1-\tfrac{1}{\rho(x)}\bigr)\bigr)\right] \\
&= KL(P_s\Vert P_t)
   - \mathbb{E}_{P_s}\!\left[\log\!\bigl(1-\beta\bigl(1-\tfrac{1}{\rho(x)}\bigr)\bigr)\right].
\end{align*}
Applying the Taylor expansion $\log(1-u)=-u+O(u^2)$, valid uniformly
under the bounded-likelihood-ratio condition~(A2)
of~\ref{app:theory:regularity}, yields
\[
KL(P_s\Vert\tilde P_t)
= KL(P_s\Vert P_t) + \beta\,\mathbb{E}_{P_s}\!\bigl[1-\tfrac{1}{\rho(x)}\bigr] + O(\beta^2).
\]
The linear coefficient simplifies via the identity
\[
\mathbb{E}_{P_s}\!\bigl[1-\tfrac{1}{\rho(x)}\bigr]
= 1 - \mathbb{E}_{P_s}\!\bigl[\tfrac{P_s(x)}{P_t(x)}\bigr]
= 1 - \!\int\! \tfrac{P_s(x)^2}{P_t(x)}\,dx
= -\mathcal{X}^2(P_s\Vert P_t),
\]
where the last equality is the definition $\mathcal{X}^2(P_s\Vert P_t) := \int P_s^2/P_t\,dx - 1$ of the Pearson $\chi^2$-divergence. This proves Lemma~\ref{lem:source_kl}. \qed

\paragraph{Proof of Lemma~\ref{lem:target_kl}.}
Write $\tilde P_t/P_t = 1 + \beta\bigl(P_s/P_t - 1\bigr)$ and let $\eta(x) := P_s(x)/P_t(x) - 1$, so
\[
KL(P_t\Vert\tilde P_t)
= -\mathbb{E}_{P_t}\!\left[\log\bigl(1+\beta\,\eta(x)\bigr)\right].
\]
Using $\log(1+v) = v - v^2/2 + O(v^3)$,
\[
KL(P_t\Vert\tilde P_t)
= -\beta\,\mathbb{E}_{P_t}[\eta(x)]
  + \tfrac{\beta^2}{2}\,\mathbb{E}_{P_t}[\eta(x)^2]
  + O(\beta^3).
\]
The first-order term vanishes because $\mathbb{E}_{P_t}[\eta(x)] = \int P_s - 1 = 0$ (both $P_s$ and $P_t$ integrate to one). The second-order term equals the Pearson $\chi^2$,
\[
\mathbb{E}_{P_t}[\eta(x)^2]
= \int \frac{\bigl(P_s(x)-P_t(x)\bigr)^2}{P_t(x)}\,dx
= \mathcal{X}^2(P_s\Vert P_t),
\]
yielding Lemma~\ref{lem:target_kl}. \qed


\subsection{Effect on Source--Target Divergence (General Case)}
\label{app:theory:source}

Fix a candidate \emph{source} distribution $P_s$ with $s\neq t$ and
assume $P_s$ shares support with $P_t$. By definition,
\begin{align*}
KL(P_s\Vert\tilde P_t)
&= \int P_s(x)\log\frac{P_s(x)}{\tilde P_t(x)}\,dx \\
&= -\mathbb{E}_{P_s}\!\left[
\log\!\left(
(1-\beta_t)\tfrac{P_t(x)}{P_s(x)}
+ \beta_t\tfrac{\bar P_{-t}(x)}{P_s(x)}
\right)\right].
\end{align*}

Introduce the likelihood ratios
\[
r_t(x) \;:=\; \frac{P_t(x)}{P_s(x)},
\qquad
\bar r(x) \;:=\; \frac{\bar P_{-t}(x)}{P_s(x)},
\]
both assumed bounded on the support of $P_s$. Factoring $r_t(x)$ out
gives
\[
\tilde P_t(x)/P_s(x)
\;=\;
r_t(x)\left(1-\beta_t\,u(x)\right),
\quad
u(x)\;:=\;1-\frac{\bar r(x)}{r_t(x)}.
\]
Substituting and splitting the logarithm,
\begin{align*}
KL(P_s\Vert\tilde P_t)
&= KL(P_s\Vert P_t)
- \mathbb{E}_{P_s}\!\bigl[\log\bigl(1-\beta_t\,u(x)\bigr)\bigr].
\end{align*}

Applying the Taylor expansion $\log(1-z)=-z-\tfrac{z^2}{2}+O(z^3)$ to
the second term (valid uniformly under the boundedness assumption,
since $|\beta_t u(x)|$ is bounded away from $1$),
\begin{equation}
KL(P_s\Vert\tilde P_t)
=
KL(P_s\Vert P_t)
+ \beta_t\,\mathbb{E}_{P_s}[u(x)]
+ \tfrac{\beta_t^{\,2}}{2}\,\mathbb{E}_{P_s}[u(x)^2]
+ O(\beta_t^{\,3}).
\label{eq:gen-source-expansion}
\end{equation}

The first-order term admits a clean information-theoretic
interpretation. Decomposing
$\bar P_{-t}=\sum_{i\neq t}\pi_{i\mid t}P_i$ and using
$\mathbb{E}_{P_s}\!\bigl[\tfrac{P_i(x)}{P_t(x)}\bigr]
= 1+\mathcal{X}^2_{(s,t)}(P_i)$, where
\[
\mathcal{X}^2_{(s,t)}(P_i)
\;:=\;
\int \frac{P_s(x)\,P_i(x)}{P_t(x)}\,dx \;-\; 1
\]
is a \emph{three-way} cross-divergence reducing to the Pearson
$\chi^2(P_s\Vert P_t)$ when $i=s$, we obtain the compact identity
\begin{equation}
\mathbb{E}_{P_s}[u(x)]
=
-\!\!\sum_{i\neq t}\!\pi_{i\mid t}\,\mathcal{X}^2_{(s,t)}(P_i).
\label{eq:gen-first-order-id}
\end{equation}

Substituting~\eqref{eq:gen-first-order-id} into
\eqref{eq:gen-source-expansion} yields the general first-order
result:
\begin{equation}
\boxed{\;
KL(P_s\Vert\tilde P_t)
=
KL(P_s\Vert P_t)
\;-\;
\beta_t\!\sum_{i\neq t}\!\pi_{i\mid t}\,
\mathcal{X}^2_{(s,t)}(P_i)
\;+\;
O(\beta_t^{\,2}).\;}
\label{eq:gen-source-main}
\end{equation}

In the single-source, single-target special case
$\bar P_{-t}=P_s$, the only nonzero term in the sum is $i=s$, for
which $\mathcal{X}^2_{(s,t)}(P_s)=\mathcal{X}^2(P_s\Vert P_t)$ by
definition, and~\eqref{eq:gen-source-main} reduces exactly to the
expansion of Lemma~\ref{lem:source_kl}.

\paragraph{Interpretation.}
Equation~\eqref{eq:gen-source-main} shows that, to first order in the
contamination strength $\beta_t$, the divergence between any source
distribution $P_s$ and the poisoned target generator decreases by a
weighted sum of cross-$\chi^2$ terms, where the weights are exactly
the components of the contamination profile $\pi_{\cdot\mid t}$.
Concretely:
\begin{enumerate}
    \item An attacker controlling a single source class $s'$
      $(\pi_{s'\mid t}=1)$ pulls the target generator toward $P_{s'}$
      and, in particular, toward any other class $P_s$ that resembles
      $P_{s'}$ (large
      $\mathcal{X}^2_{(s,t)}(P_{s'})$).
    \item Spreading the contamination across several source classes
      yields a first-order effect equal to the
      $\pi_{\cdot\mid t}$-weighted average of the corresponding cross
      terms, so distributing the attack neither amplifies nor
      cancels the leading-order distortion --- it interpolates it.
    \item Oversampling enters only through $\beta_t$: increasing the
      reweighting factors $r_{ij}$ inflates the $\Delta_{ij}$
      in~\eqref{eq:gen-delta}, hence $\beta_t$, without altering
      the contamination profile if $R$ is proportional across $i$.
\end{enumerate}

\subsection{Effect on Target Distribution (General Case)}
\label{app:theory:target}

By definition,
\begin{align*}
KL(P_t\Vert\tilde P_t)
&= -\mathbb{E}_{P_t}\!\left[
\log\!\left(1+\beta_t\!\left(\tfrac{\bar P_{-t}(x)}{P_t(x)}-1\right)\right)
\right].
\end{align*}
Let $v(x)\;:=\;\tfrac{\bar P_{-t}(x)}{P_t(x)}-1$, and note that
$\mathbb{E}_{P_t}[v(x)]=\int \bar P_{-t}-1 = 0$ since $\bar P_{-t}$
integrates to $1$. Using
$\log(1+z)=z-\tfrac{z^2}{2}+O(z^3)$ (valid under bounded ratios),
\begin{align*}
KL(P_t\Vert\tilde P_t)
&= -\beta_t\,\mathbb{E}_{P_t}[v(x)]
+ \tfrac{\beta_t^{\,2}}{2}\,\mathbb{E}_{P_t}[v(x)^2]
+ O(\beta_t^{\,3}) \\
&= \tfrac{\beta_t^{\,2}}{2}\,\mathbb{E}_{P_t}[v(x)^2]
+ O(\beta_t^{\,3}).
\end{align*}
The variance term is exactly the Pearson $\chi^2$-divergence of the
\emph{aggregate} contaminating distribution against the true target,
\[
\mathbb{E}_{P_t}[v(x)^2]
=
\int \frac{\bigl(\bar P_{-t}(x)-P_t(x)\bigr)^2}{P_t(x)}\,dx
=
\mathcal{X}^2\bigl(\bar P_{-t}\Vert P_t\bigr).
\]
We therefore obtain the general second-order result:
\begin{equation}
\boxed{\;
KL(P_t\Vert\tilde P_t)
=
\frac{\beta_t^{\,2}}{2}\,
\mathcal{X}^2\!\left(\bar P_{-t}\,\big\Vert\,P_t\right)
+ O(\beta_t^{\,3}).\;}
\label{eq:gen-target-main}
\end{equation}

In the single source--target case $\bar P_{-t}=P_s$,
Eq.~\eqref{eq:gen-target-main} collapses to the expansion of
Lemma~\ref{lem:target_kl}.

\paragraph{Interpretation.}
The deviation from the true target distribution is governed entirely
by the \emph{aggregate} contaminating distribution $\bar P_{-t}$,
regardless of how that aggregate is composed across individual source
classes. Two distinct attack policies that yield the same
$\bar P_{-t}$ and $\beta_t$ are second-order indistinguishable from
the target marginal alone. By the convexity of
$\mathcal{X}^2(\cdot\Vert P_t)$ in its first argument,
\[
\mathcal{X}^2(\bar P_{-t}\Vert P_t)
\;\le\;
\sum_{i\neq t}\pi_{i\mid t}\,\mathcal{X}^2(P_i\Vert P_t),
\]
so \emph{spreading} the attack across multiple source classes can
\emph{reduce} the second-order signal $KL(P_t\Vert\tilde P_t)$ at the
\emph{same} total contamination $\beta_t$. This refines the
first/second-order asymmetry observed in
Section~\ref{sec:theoretical}: a sophisticated attacker can not only
exploit the inherent $\beta_t$ vs.\ $\beta_t^2$ gap, but additionally
flatten $\bar P_{-t}$ across several source classes to suppress
detection-relevant statistics while preserving the linear damage to
$KL(P_s\Vert\tilde P_t)$.

\subsection{Regularity Assumptions}
\label{app:theory:regularity}

The expansions~\eqref{eq:gen-source-expansion}
and~\eqref{eq:gen-target-main} are justified under the following
mild conditions, which are direct generalizations of the bounded
likelihood-ratio condition of Section~\ref{sec:theoretical}:
\begin{description}
    \item[\textnormal{(A1)}] \textit{Shared support.}
      For every $i\in\mathcal{C}$ with $\pi_{i\mid t}>0$,
      $\mathrm{supp}(P_i)\subseteq\mathrm{supp}(P_t)$, and likewise
      for $P_s$ relative to $P_t$.
    \item[\textnormal{(A2)}] \textit{Bounded likelihood ratios.}
      There exist constants $0<m\le M<\infty$ such that
      \[
      m \;\le\; \frac{P_i(x)}{P_t(x)} \;\le\; M
      \quad\text{for all } x\in\mathrm{supp}(P_t),\; i\in\mathcal{C}.
      \]
    \item[\textnormal{(A3)}] \textit{Non-degenerate contamination.}
      $\beta_t < 1$, i.e., honest clients contribute strictly positive
      mass to the training pool for class $t$.
\end{description}
Under (A1)--(A2), all $\chi^2$-type quantities appearing
in~\eqref{eq:gen-source-main} and~\eqref{eq:gen-target-main} are
finite, and the Taylor remainder terms in the expansions are
uniformly $O(\beta_t^{\,2})$ and $O(\beta_t^{\,3})$ respectively.
Assumption (A3) ensures the mixture~\eqref{eq:gen-mix-compact} is
well-defined and that the expansions are taken about an interior
point of $[0,1]$.

\subsection{Special Cases}
\label{app:theory:special}

We highlight three regimes that follow as immediate corollaries of
Eqs.~\eqref{eq:gen-source-main}--\eqref{eq:gen-target-main}.

\paragraph{(i) Single source, single target.}
Taking $Q$ to flip only $s\to t$ with probability $p$ and $R$ to
oversample only that pair by $r$ recovers $\beta_t=\beta$ and
$\bar P_{-t}=P_s$, giving exactly the expansions of
Lemmas~\ref{lem:source_kl} and~\ref{lem:target_kl}.

\paragraph{(ii) Uniform contamination from $\mathcal{S}\subset\mathcal{C}\setminus\{t\}$.}
If the attacker flips a subset $\mathcal{S}$ uniformly into $t$ with
equal probability and equal oversampling, then
$\pi_{i\mid t}=\mathbf{1}\{i\in\mathcal{S}\}/|\mathcal{S}|$ and
$\bar P_{-t}=|\mathcal{S}|^{-1}\sum_{i\in\mathcal{S}}P_i$. The
first-order damage~\eqref{eq:gen-source-main} averages the
$\mathcal{X}^2_{(s,t)}(P_i)$ over $i\in\mathcal{S}$, while the
second-order signal~\eqref{eq:gen-target-main} is governed by
$\mathcal{X}^2$ of the \emph{averaged} contaminant against $P_t$, which
is generally strictly smaller than the average of the $\chi^2$'s
(Jensen).

\paragraph{(iii) Targeting multiple classes simultaneously.}
If the attacker poisons several target classes $t\in\mathcal{T}$, the
analysis decouples class-by-class: each
$\tilde P_t$ for $t\in\mathcal{T}$ admits
its own $(\beta_t,\bar P_{-t})$, and~\eqref{eq:gen-source-main}--\eqref{eq:gen-target-main}
hold independently for each $t$. The single source--target setting
analyzed in the main text is thus a building block from which
arbitrary confusion-matrix attacks can be assembled and analyzed
additively at leading order.
\end{document}